# Perception of visual numerosity in humans and machines

Alberto Testolin[1,*], Serena Dolfi[1], Mathijs Rochus[2], Marco Zorzi[1,3,*]

[1] Department of General Psychology and Padova Neuroscience Center,
University of Padova, 35131 Padova, Italy

[2] Department of Experimental Psychology, Ghent University, 9000 Ghent, Belgium

[3] IRCCS San Camillo Hospital, 30126 Venice-Lido, Italy

* Correspondence: alberto.testolin@unipd.it or marco.zorzi@unipd.it

**Abstract**

Numerosity perception is foundational to mathematical learning, but its computational bases are strongly debated. Some investigators argue that humans are endowed with a specialized system supporting numerical representation; others argue that visual numerosity is estimated using continuous magnitudes, such as density or area, which usually co-vary with number. Here we reconcile these contrasting perspectives by testing deep networks on the same numerosity comparison task that was administered to humans, using a stimulus space that allows to measure the contribution of non-numerical features. Our model accurately simulated the psychophysics of numerosity perception and the associated developmental changes: discrimination was driven by numerosity information, but non-numerical features had a significant impact, especially early during development. Representational similarity analysis further highlighted that both numerosity and continuous magnitudes were spontaneously encoded even when no task had to be carried out, demonstrating that  numerosity is a major, salient property of our visual environment.

It is widely believed that the cognitive foundations of mathematical learning rest on basic numerical intuitions, such as the ability to discriminate sets with different numerosity or to rapidly estimate the amount of items in a display (1–4). Discrimination between two numerosities is modulated by their numerical ratio (5) and the individual psychometric function provides an index of "number acuity" (the internal Weber fraction, $w$; (6)). Numerosity perception is shared with many animal species (7–10) and in the primate brain it is supported by an occipito-parietal network (6, 11, 12). Even newborns and infants appear sensitive to numerosity (13, 14), although the improvement of number acuity throughout childhood suggests that learning and sensory experience play an important role in refining our numerical representations (15, 16). Moreover, individual differences in number acuity have been related to mathematical learning performance both in typical and atypical development (4, 17–19).

Nevertheless, the nature of the mechanisms underlying numerosity perception is still hotly debated. According to the "number sense" hypothesis, visual numerosity is a primary perceptual attribute (20), spontaneously extracted (21, 22) by a system yielding an approximate representation of numerical quantity (23). However, it has been repeatedly pointed out that numerosity judgments can be modulated by non-numerical perceptual cues that usually co-vary with number, such as cumulative surface area (24, 25), total item perimeter (26) and convex hull (27), over which it is impossible to exert full experimental control (28, 29) and which can hinder numerosity discrimination when carrying incongruent information (30–32). These findings have led to the proposal that numerosity is indirectly estimated from non-numerical visual features, thereby calling into question the existence of a dedicated system for numerosity perception (33, 34). Moreover, the influence of non-numerical cues is stronger in young children (35) and in children with mathematical learning deficit (36, 37). In a classic "number sense" view, the developmental improvement of numerosity estimation entails progressive sharpening of the internal representation (i.e., increasing representational precision), but an alternative hypothesis is that it simply reflects the increasing ability to focus on the relevant dimension and filter out (or inhibit) irrelevant non-numerical features (31, 36).

Here we shed light on the theoretical debate regarding the mechanism supporting numerosity perception and how it is shaped by learning and visual experience by means of computational modeling based on deep neural networks. Besides driving the contemporary artificial intelligence revolution (38), deep networks are being increasingly proposed as models of neural information processing in the brain (39–43). In contrast to the mainstream supervised deep learning approach, which has been criticized for its limited biological and psychological plausibility (44, 45), our computational model is based on deep belief networks, which implement unsupervised learning through Hebbian-like principles (46). In line with modern theories of cortical functioning, learning in these networks can be interpreted as the process of fitting a probabilistic, generative model to the sensory data (47, 48). This dispenses from the implausible assumption that learning requires labeled examples, since the only objective is to discover meaningful internal representations of the environment (45).

The importance of unsupervised representation learning is particularly evident in the context of numerosity perception, which develops in infants and naive animals without explicit feedback (49, 50). Previous work has shown that hierarchical generative models can simulate basic numerical abilities (51, 52): here we push this modeling approach one step further, by adopting a recently proposed methodology that allows teasing apart the effects of non-numerical features in numerosity perception (53). Crucially, we test human observers and deep networks on a numerosity comparison task using exactly the same visual stimuli, thus guaranteeing perfect overlap of experimental conditions and allowing for accurate simulations of psychophysical data. We compare deep networks at different time points during unsupervised learning to assess how visual experience shapes the interplay between numerosity and non-numerical visual features. Furthermore, we use representational similarity analysis (54) to investigate the spontaneous sensitivity to number and/or continuous visual features in the networks' internal representations regardless of the numerosity comparison task.

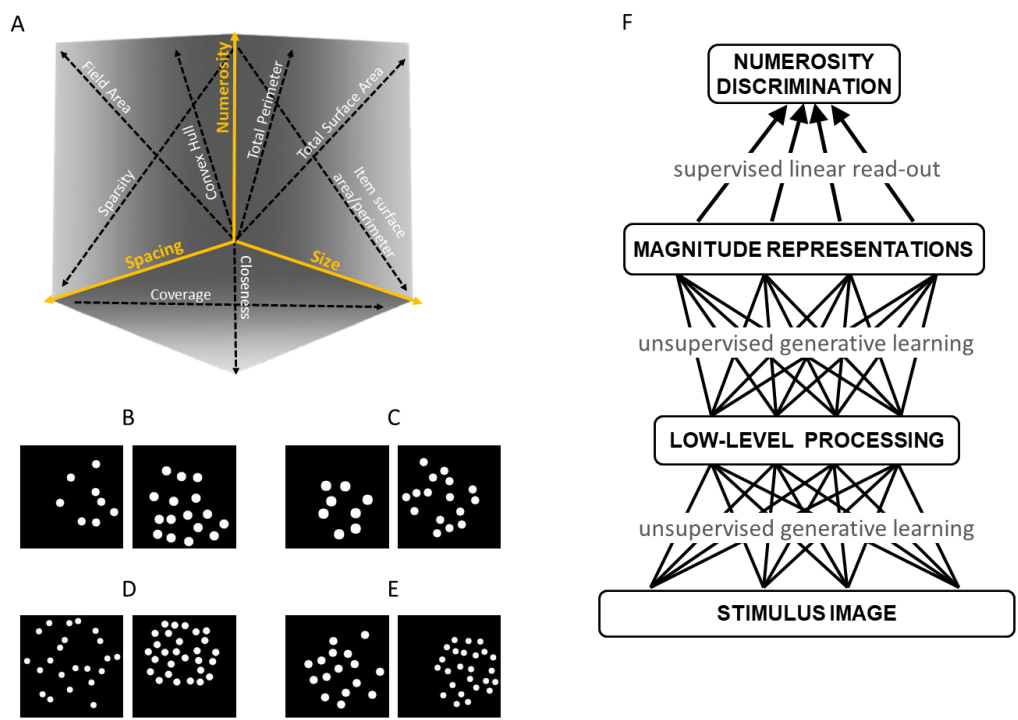

**Figure 1. Stimulus space and model architecture**. (A) The 3D stimulus space defined by the Numerosity, Size and Spacing orthogonal dimensions (adapted from (53)). Non-numerical features are represented as arrows to indicate the direction in which they increase, and each stimulus image can be represented as a point in this space. Example of stimuli pairs are shown below, where Numerosity can be fully congruent for Size and Spacing (B), congruent for Spacing but not for Size (C), congruent for Size but not for Spacing (D), or fully incongruent for Size and Spacing (E). The model architecture is depicted in panel (F). At the initial stage, unsupervised deep learning adapts the connection weights of the first two layers (undirected edges) by capturing the statistical distribution of active pixels in the images. During task learning, a supervised linear classifier adapts the connection weights of the final layer (directed edges) in order to minimize discrimination error.

# Results

Image stimuli containing clouds of dots were sampled from a multidimensional space defined by three orthogonal dimensions, representing the degrees of freedom used to generate all possible combinations of numerosity and non-numerical features in a visual display (53). *Numerosity* corresponds to the discrete number of dots in the image, *Spacing* jointly encodes for variations in field area and density of the dots, while *Size* jointly encodes for variations in dot surface area and total surface area (see Fig. 1A-E for graphical representation and sample stimuli, and Supplementary Information for details). Humans and machines were probed using a standard numerosity comparison task that required to indicate which of two images had more dots. Behavioral choices were modeled using a generalized linear model (GLM) with regressors for the log of the ratio of each orthogonal dimension (see Methods). Besides offering a better estimate of number acuity compared to traditional measures (53), this method returns coefficients describing the contribution of each non-numerical feature in behavioral performance. For example, a large Numerosity coefficient would reflect the ability to discriminate difficult numerical ratios, while large coefficients for Size and Spacing would highlight strong biases on the subject's choices due to changes in non-numerical features.

## *Human Behavioral Performance*

Discrimination accuracy was well above chance (mean 83%, range: 69 – 91%). The GLM fit at individual level was significant (mean adjusted $R^2$ = .55, mean chi-square value = 191.14, all $p$ < .001). Coefficient fits for each orthogonal dimension were significantly different from zero for *βNum* ($t$(39) = 23.54, $p$ < .001) and *βSpacing* ($t$(39) = 7.21, $p$ < .001), but not for *βSize* ($t$(39) = 1.37, $p$ = .18). Coefficient estimates are shown in the scatter plots of Fig. 2A, along with the axes representing individual features. Numerosity was by far the dominant dimension in shaping participants' choices. Nevertheless, most of the participants were also influenced by Size, Spacing, or both; only 5 subjects out of 40 showed an unbiased performance (all $ts$ < 1.35, $p$ > .10 for *βSize*, and all $ts$ < 1.45, $p$ > .15 for *βSpacing*). Model fits for two representative participants are shown in Fig. S1 (panels A-B), with black curves representing the model fit on the full dataset and colored lines representing model predictions on two subsets of congruent and incongruent trials with extreme values of Size and Spacing ratios. The offsets of the colored lines from the black lines highlight that these participants were influenced by Size (panel A) or both Size and Spacing (panel B), resulting in a better discrimination for congruent trials and a decrease in performance for incongruent trials. The psychometric curves also highlight that, as expected, accuracy was much lower for hard trials (i.e., with ratio closer to 1, or log ratio closer to 0) compared to easy trials.

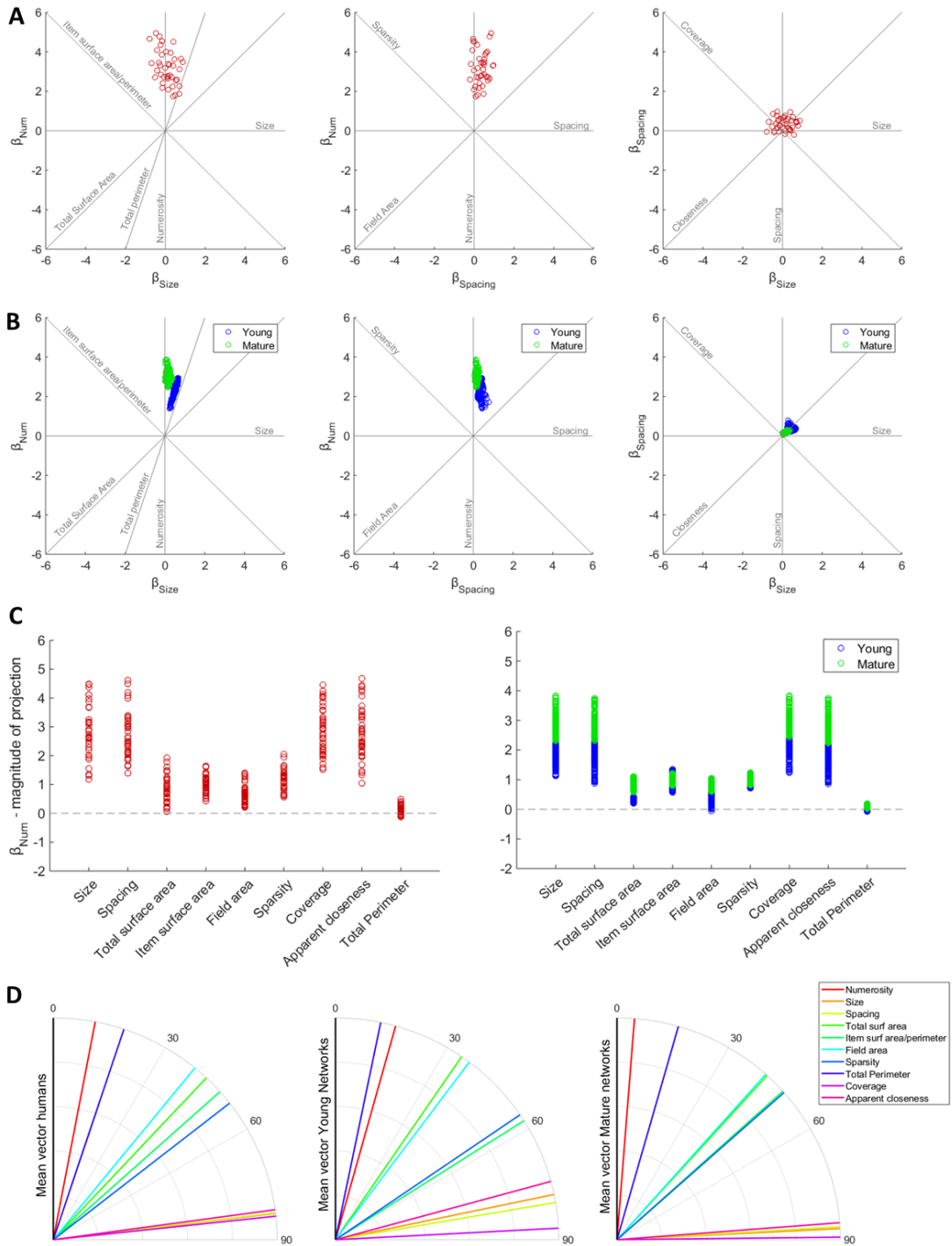

**Figure 2. Psychophysics of numerosity comparison in humans and deep networks**. Scatter plots of Numerosity, Size and Spacing coefficients for humans (A) and all deep networks (B). (C) Differences between the projection on the Numerosity dimension and the projections on all individual non-numerical features for humans (left) and deep networks (right). Positive values indicate that number was a better predictor of behavior than the specific feature. Negative values would indicate that the considered feature had greater impact on discrimination choice. (D) Angles between the discrimination vector and all non-numerical features (the discrimination vector is on the *y* axis) for humans (left) and deep networks at two different developmental stages: Young (middle) and Mature (right).

In the 3-dimensional space, the coefficients estimated for the orthogonal dimensions define a discrimination vector whose projection on each individual axis indicates the strength of the influence of the corresponding feature (left panel in Fig. 2C). Paired t-tests revealed that the projection on the Numerosity dimension was significantly larger than the projection on all other individual features (all $t(39) > 5.45$, all $p < .001$), replicating the findings of DeWind et al (2015). The overall pattern was confirmed by angle analysis (left panel in Fig. 2D), in which Numerosity resulted the dimension closest to the discrimination vector (10.79 deg) followed by Total Perimeter (18.39 deg); the angle between these two axes was significantly different ($Z = 4.14$, $p < .001$).

*Deep Networks Behavioral Performance*

The same numerosity comparison task was simulated with deep neural networks. Deep networks (N = 144) with varying architectures and initial weights (see Methods and SI) were first trained in a completely unsupervised way, using as input individual images sampled from the stimulus space. The objective of unsupervised deep learning was to build a generative model of the data, that is, to maximize the likelihood of reproducing samples from the input distribution. This corresponds to a form of learning by observation, where connection weights are changed according to the statistical regularities in the sensory input; no information about item numerosity was provided at this stage. As a second step, a linear network was stacked on top of the deep network to simulate task learning. This classifier was fed with the deep network's internal representations of two image pairs, with the objective of choosing which image had the larger numerosity (see Fig. 1F for a graphical representation of the model architecture). Testing was carried out on a separate set of images that were never seen during training and it took place at two different developmental points of the unsupervised learning phase. In the "Young" condition the network was trained for only one pass (epoch) through the entire image dataset, whereas in the "Mature" condition training was prolonged for 200 epochs[1].

As shown in Fig. 2B, the GLM coefficients for the 144 individual deep networks were aligned with those of human observers, highlighting the primary contribution of Numerosity but also the impact of Size and Spacing in biasing the numerosity judgments. All coefficient fits were significantly different from zero both for Young networks ($\beta Num$ all $t > 66$, $p < .001$; $\beta Spacing$ all $t > 17.9$, $p < .001$; $\beta Size$ all $t > 32$, $p < .001$) and Mature networks ($\beta Num$ all $t > 66$, $p < .001$; $\beta Spacing$ all $t > 6.9$, $p < .001$; $\beta Size$ all $t > 2.6$, $p < .001$)[2]. It is important to emphasize that variability across networks was limited, thereby showing that the modeling results are robust to changes in architecture and initial state. GLM fits for representative Young and Mature networks are shown in Fig. S1 (panels C-D); as for human participants, the offsets of the colored lines from the black lines highlight that model choices were influenced by Size and Spacing, resulting in better discrimination for congruent trials and worse discrimination for incongruent trials, especially for Young networks.

---

[1] We should note that the choice of 1 *vs.* 200 epochs is quite arbitrary and does not imply a particular correspondence with human developmental stages. Indeed, we will mostly focus on comparing the *change* over development, rather than absolute values at each developmental stage.

[2] Note that the Numerosity coefficient can be directly related to a more classic measure of number acuity, that is, the internal Weber fraction ($w = 1/(\sqrt{2} * \beta Num)$, see (53)). Thus, the change in $\beta Num$ across Young and Mature networks would correspond to an increase in number acuity from 0.32 to 0.22, a range compatible with that observed during childhood (15, 19).

As shown in Fig. 2C (right panel), for Young networks the vector projection on the Numerosity dimension was larger than the projections on the other individual features (all $Z > 10.40$, all $p < .001$) except for Total Perimeter, whose projection resulted higher than $\beta Num$ ($Z = -10.41$, $p < .001$). For the Mature networks, the vector projection on Numerosity was larger than all other projections (all $Z > 10.41$, all $p < .001$). This pattern was confirmed by angle analysis (Fig. 2D): in the Mature networks Numerosity was the dimension closest to the discrimination vector (4.49 deg) followed by Total Perimeter (15.96 deg), and the angle between these two dimensions was significantly different ($Z = 10.41$, $p < .001$). However, in the Young networks the closest dimension was Total Perimeter (11.69 deg), followed by Numerosity (15.61 deg), and the angle between these two dimensions was significantly different ($Z = -10.41$, $p < .001$).

Unsupervised learning thus seems to have strengthened the numerical information encoded in the deep networks, thereby supporting higher accuracy in the numerosity comparison task. As shown in Fig. 3A, $\beta Num$ was much higher for the Mature network compared to the Young network ($U = 713$, $p < .001$), while the influence of both Size ($U = 20535$, $p < .001$) and Spacing ($U = 20393$, $p < .001$) significantly decreased. At the level of individual features (Fig. 3B) we also observed a large difference in angles with the discrimination vector: the angle with Numerosity significantly decreased in the Mature network ($U = 20736$, $p < .001$), while there was a significant increase of the angles with Total Perimeter ($U = 2623$, $p < .001$) and Total Surface Area ($U = 142$, $p < .001$). These results are well aligned with recent human developmental data collected using the same stimulus space (35): the comparison between 4-year-old children (the youngest group tested) and adults reveals a marked increase in the Numerosity coefficient and a moderate decrease of Size and Spacing coefficients, as well as a significant reduction in angle with Numerosity (see Fig. 3C). This developmental change is very similar to that observed in the deep networks between Young and Mature states. Overall, these findings suggest that numerosity is the primary feature driving numerical comparison both in children and adults, but with a different non-numerical bias that depends on the amount of sensory experience (which affects the quality of the learned generative model).

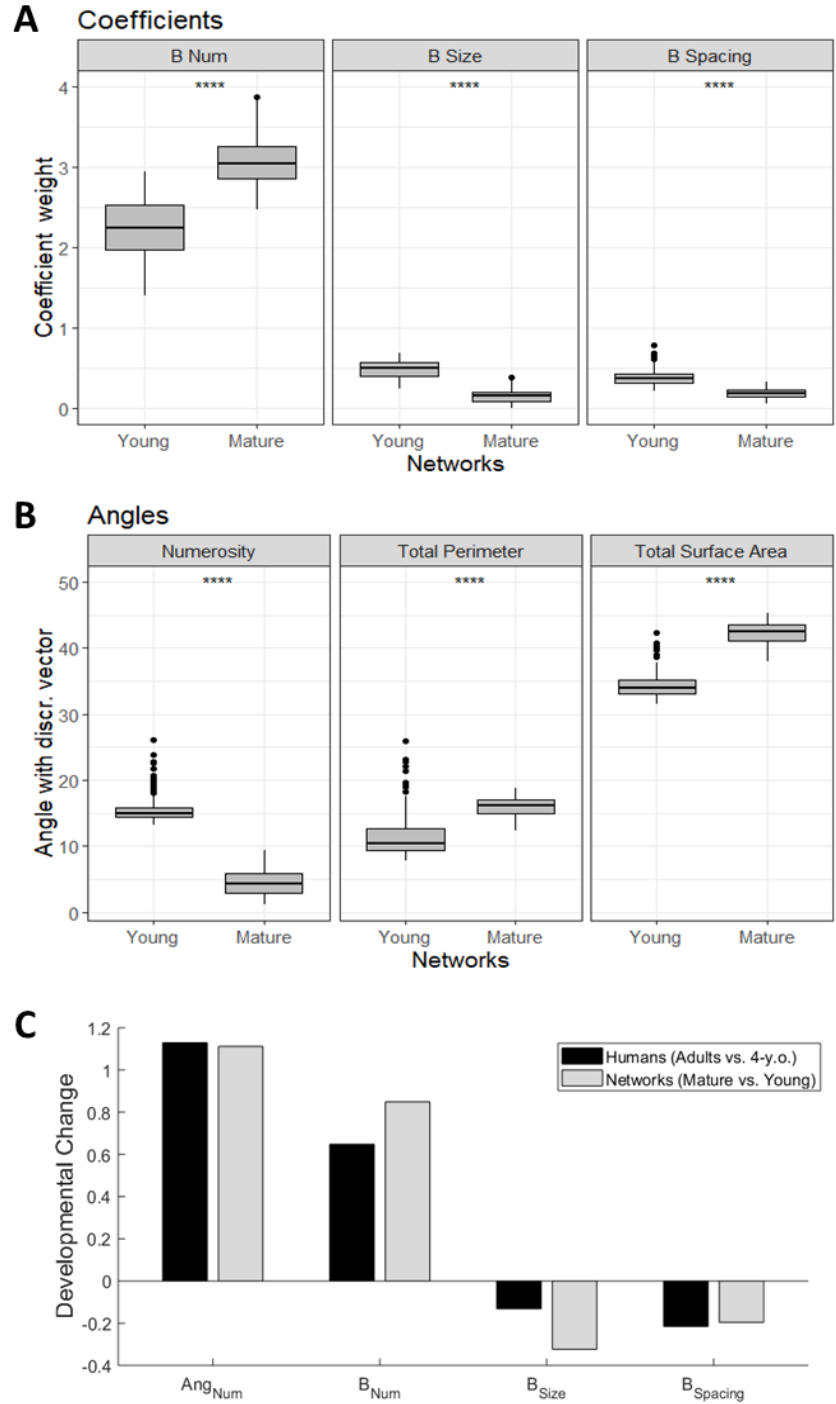

**Figure 3. Maturation of number acuity in deep networks**. (A) Differences in Numerosity, Size and Spacing coefficients, measured separately for the Young and Mature networks. (B) Differences in angles between the discrimination vector and the most relevant non-numerical features, measured separately for the Young and Mature networks. (C) Comparison between angle and coefficients changes in humans (data replotted from Starr et al., 2017) and deep networks. Note that angle differences for both humans and networks have been scaled by a factor of 10 for visualization purposes.

*Deep Networks Internal Encoding*

We systematically analyzed the deep networks' internal representations to disentangle the contribution of unsupervised deep learning in numerosity perception from the (supervised) task-driven decoding of information supporting numerosity judgments. To this aim we investigated how numerosity and non-numerical features were spontaneously encoded in the activation patterns of hidden neurons in response to individual images, in the absence of an explicit task that requires focusing on numerical information as a salient dimension to guide overt behavior (i.e., without considering the discrimination layer).

We performed a Representational Similarity Analysis (RSA) (54) to assess which features were encoded in the network's internal representations. We compared the Representational Dissimilarity Matrix (RDM) obtained from the best performing deep network architecture with conceptual RDMs reflecting specific categorical models (Fig. 4A). The correlations between simulated RDMs and categorical models can be visualized as a second-order correlation matrix (Fig. 4B). The Young network exhibited a stronger correlation (Kendall Tau alpha, see Fig. 4C) with the RDM produced using Convex Hull ($\tau_A = .69$), followed by Total Perimeter ($\tau_A = .52$), Numerosity ($\tau_A = .51$) and Field Area ($\tau_A = .47$); correlation with all other categorical models was smaller, but significant in a one-sided signed rank test, thresholded at FDR < .01 (55). The Mature network, instead, had stronger correlations with the RDMs for Numerosity ($\tau_A = .33$), Total Perimeter ($\tau_A = .33$) and Convex Hull ($\tau_A = .30$), and significant correlations also with all the other categorical models (FDR < .01). Although the correlation with Numerosity in fact decreased during learning, the increase in correlation with all categorical models suggests that the representational space become more disentangled, thus allowing for a better factorization of the latent features of the stimulus space. Pairwise comparisons are shown in Fig. S2, highlighting the primary role of Convex Hull during early developmental stages, but an increase in the contribution of Numerosity later in development.

Overall, RSA showed that numerosity information was spontaneously encoded as a salient dimension in the representational geometry of the deep network, even when no explicit numerosity judgements had to be carried out. A compatible result was obtained using t-SNE (56), which is a technique commonly used for visualizing the high-dimensional representational space of deep networks (see Fig. S3 and SI). When t-SNE was performed on the internal representations of images where Numerosity, Size and Spacing were all congruent, the algorithm was able to project patterns corresponding to small and large magnitudes into clearly separated clusters. When either Size or Spacing information was incongruent with number, the separation was still possible, but more evident in Mature networks. When both Size and Spacing were incongruent with number, the internal representations of Young networks did not support the formation of separate clusters.

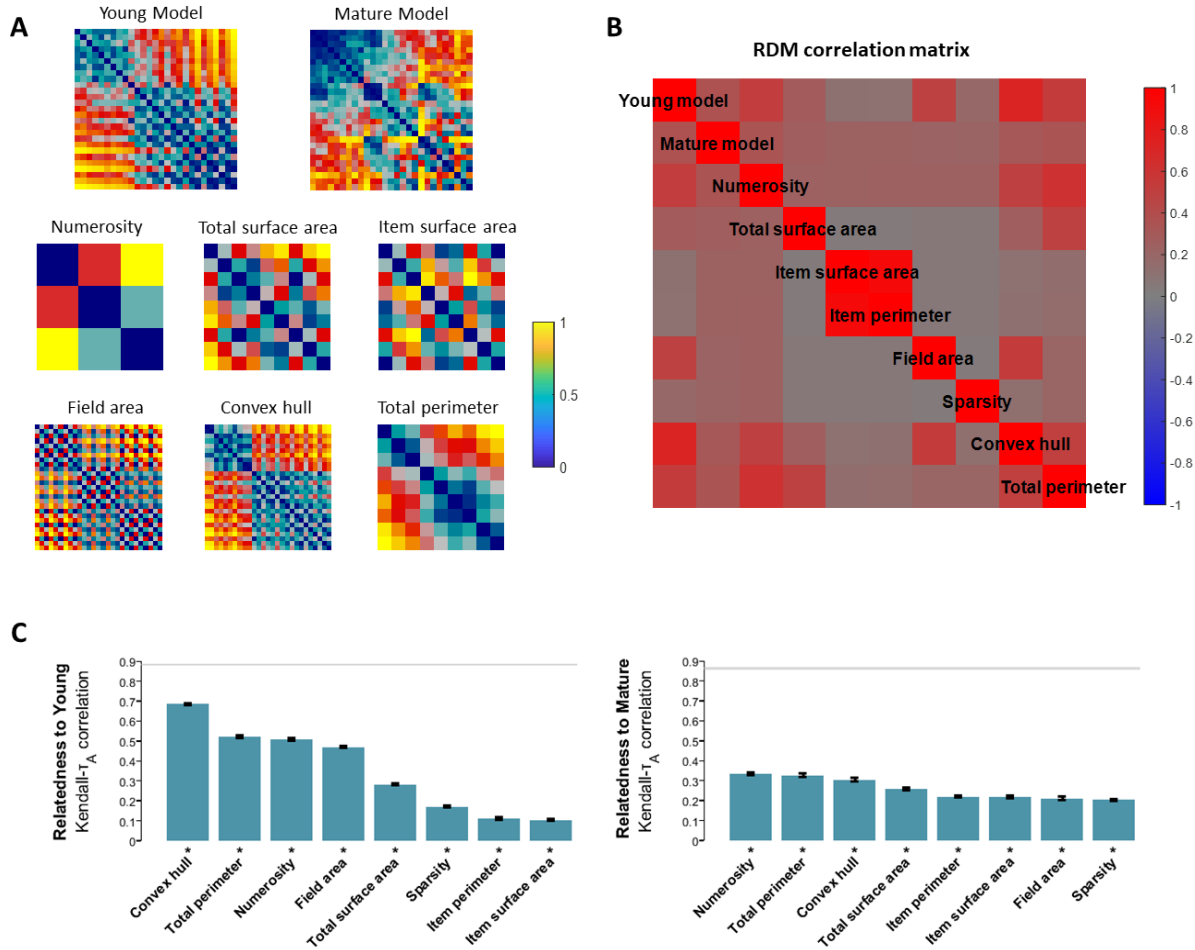

**Figure 4. Representational similarity analysis.** (A) Representational dissimilarity matrices for the best deep network architecture (distance measure: 1 – Pearson correlation) and the most relevant categorical models (distance measure: log distance between stimulus features). Each RDM was separately rank transformed and scaled into [0,1]. (B) Second-order correlation matrix showing the pairwise correlations between RDMs. (C) Relatedness between the model's RDM and the categorical RDMs, measured as the Kendall rank correlation between dissimilarity matrices. Asterisks indicate significance in a one-sided signed rank test, thresholded at FDR < 0.01. Error bars indicate the standard error of the correlation estimate. Grey horizontal lines represent noise ceiling (i.e., the highest correlation that could be achieved considering the data variability).

## Discussion

The computational investigations presented in this article reconcile two contrasting perspectives about the nature of our visual number sense. On the one hand, numerosity turned out to be the primary driver of both humans' and deep networks' responses in a numerosity comparison task, even when non-numerical visual cues were included as predictors of behavioral choices. Numerosity was also a critical factor in shaping the internal representations emerging from unsupervised deep learning, showing that number was spontaneously encoded in the model even when no number-related decision had to be carried out (and indeed, in the absence of any task). These results support the characterization of numerosity as a primary perceptual attribute (22) and are aligned with the hypothesis that animals are endowed with a number sense (1, 3). On the other hand, continuous visual features had a significant impact both in humans' and deep networks' responses, thus confirming that numerosity estimation is modulated by non-numerical magnitudes that usually co-vary with number (33, 34). Interestingly, many of these alternative dimensions were also spontaneously encoded in the representational space of the model, suggesting that non-numerical features are equally important in order to capture the latent factors of variation underlying the sensory data (57). In this respect, the influence of non-numerical visual features might thus be seen as the concurrent processing of other dimensions carrying magnitude information, but without necessarily implying that numerosity is constructed out of those dimensions.

Our simulations show that numerosity processing can be carried out using generic low-level computations, such as those emerging in multi-layered neural networks that learn a hierarchical generative model of the sensory data. Previous modeling work has shown that high-frequency spatial filtering of the image is the building block for computing numerical information at higher levels in the neural processing hierarchy (51, 52, 58). This is consistent with a psychophysical modeling approach based on spatial filtering (58) and with recent electrophysiological evidence suggesting that numerosity-sensitive responses are present at early stages of visual processing (12, 59). Numerical information in the deep neural network is best conceived as a population code over the hidden neurons (51, 52), as also shown here by the representational similarity analysis. Though the latter results do not necessarily imply the existence of a dedicated system for processing numerosity, sensitivity to numerosity has been found even at the level of individual hidden neurons (51, 52), with tuning functions resembling those of real neurons in the dorsal visual stream of non-human primates (59). However, it is worth noting that the modulatory effect of non-numerical visual features awaits systematic examination in neurophysiological studies.

The comparison between Young and Mature deep networks showed an overall improvement in number acuity, in line with developmental studies. This improvement reflected the increased weighting of numerical information and the concurrent down-weighting of the non-numerical dimensions (35). Such developmental change might be interpreted in terms of an improved ability to focus on numerosity and to filter out task-irrelevant features, so that the discrimination boundary gets progressively aligned to the task-relevant dimension (36). However, in our simulations the discrimination layer received identical training when applied to both Young and Mature networks, thereby suggesting that the improvement stems from a refinement of the internal representations following unsupervised learning. In other words, a sharper internal encoding allows a better disentanglement of numerosity from other dimensions defining the statistical structure of the visual environment. The sharpening hypothesis is also supported by our representational similarity analysis, which shows that numerosity is indeed encoded even in the absence of a task, and it becomes better factorized in the Mature state.

The finding that supervised deep networks trained in visual discrimination tasks may show idiosyncratic (non-human) behavior (60) has raised concerns about their capacity to faithfully mimic human vision. Here we observed an impressive match between human performance and deep neural networks, which suggests that neurocomputational models based on unsupervised deep learning represent a powerful framework to investigate the emergence of perceptual and cognitive abilities in learning machines that emulate the processing mechanisms of real brains (also see (40)). Furthermore, our exploration of different architectures and learning hyperparameters suggests that numerosity comparison is a challenging task for deep learning models, although the present results do not exclude the possibility that more advanced architectures (e.g., incorporating *ad-hoc* pre-processing stages or convolutional mechanisms) could achieve higher performance[3]. The response variability exhibited by the different deep learning architectures also suggests that this framework could be used to study the factors contributing to the emergence of individual differences in human observers, which is crucial for developing personalized computational models that may predict mathematical learning outcomes (see (61) for a recent application to learning to read and dyslexia).

Besides improving our current understanding of the computational foundations of numerosity perception, our modeling work has also the potential for technological applications. For example, intelligent machines that can perceive and manipulate numerosity in a meaningful way would allow to replace human annotators in tedious tasks, such as estimating the number of cells in microscopic images (62), monitoring crowds and traffic congestion in automatic surveillance systems (63, 64) or the number of trees in aerial images of forests (65). Number-related questions are also being included in standard benchmarks for assessing intelligent dialogue systems (66), but the ability to flexibly manipulate numbers and perform quantitative reasoning is still out of reach even for state-of-the-art systems (67). Indeed, although computers largely outperform humans on tasks requiring the mere application of syntactic manipulations (e.g., performing algebraic operations on large numbers, or iteratively computing the value of a function), they completely lack a conceptual semantics of number. The problem of grounding symbolic knowledge into some form of intrinsic meaning is well known in artificial intelligence research (68, 69), and mathematics constitutes one of the most challenging domains for investigating how abstract symbolic notations could be linked to bottom-up, sensorimotor percepts (70). We believe that our modelling work constitutes an important step towards a better understanding of the mechanisms underlying our visual number sense, paving the way for the development of a computational theory of mathematical learning and its impairments in atypical populations.

---

[3] Note, however, that even state-of-the-art models, such as those base on generative adversarial networks (76), fall short in representing numerosity as a disentangled factor (77).

## Methods

*Human participants.* Forty volunteer students (mean age 23.7 years, range 20 – 28, 32 females) were recruited at the University of Padova. All participants gave written informed consent to the protocol approved by the Psychological Science Ethics Committee of the University of Padova and did not receive any payment.

*Visual Stimuli.* Images of size 200x200 pixels were generated by randomly placing white dots on a black background. For the discrimination task there were 13 levels of Numerosity (range 7 – 28), 13 levels of Size (range 2.6 – 10.4 pixels x$10^5$) and 13 levels of Spacing (range 80 – 320 pixels x$10^5$), evenly spaced on a logarithmic scale. For each selected point in the stimulus space 10 different images were generated by randomly varying dots displacement, resulting in a dataset of 21970 unique images. For the human experiment we randomly selected images from the dataset to create 300 image pairs with different magnitude ratios, oversampling the more difficult numerosity ratios (10% with ratio between 0.5 and 0.6; 20% with ratio between 0.6 and 0.7; 30% with ratio between 0.7 and 0.8; 40% with ratio between 0.8 and 0.9). For simulations, 15200 image pairs were created by randomly choosing among all patterns in the dataset. We also created an independent dataset of 65912 images, containing all numbers between 5 and 32, which was used only for unsupervised learning.

*Procedure for the human study.* Stimuli were projected on a 19-inches color screen. Participants sat approximately 70 cm from the screen and placed their head on a chin rest. Subjects were verbally instructed to select the stimulus with more dots, responding with the left and right arrows of the keyboard depending on its side of appearance (feedback was given only during few practice trials). The task consisted in 3 blocks of 100 trials each, for a total of 300 trials. Each trial began with a fixation cross at the center of the screen (500 ms), followed by the simultaneous presentation of two stimuli (250 ms), one at the right and one at the left of the cross with eccentricity of ~12 visual degrees, and then by two masks of black and white Gaussian noise in the same positions (150 ms). A black screen was then displayed until response, without time limit. After response, a pseudorandom inter-trial interval between 1250 and 1750 ms occurred. Subjects also performed a sequential version of the same task, whose outcome was aligned with the simultaneous version and thus it is not further considered (see SI).

*GLM Analysis.* All responses below stimulus presentation time were considered outliers, as well as response times over two standard deviations from the subject's mean response time in equally difficult trials (based on numerosity ratio). A generalized linear model was then fitted to the choice data of each subject (53), which was modeled as a function of the three regressors Numerosity, Size and Spacing:

$$p(ChooseRight) = (1-\gamma)\left(\frac{1}{2}\left[1 + erf\left(\frac{\beta_{Side} + \beta_{Num}\log_2(r_{num}) + \beta_{Size}\log_2(r_{size}) + \beta_{Spacing}\log_2(r_{Spacing})}{\sqrt{2}}\right)\right] - \frac{1}{2}\right) + \frac{1}{2}$$

where $r_{num}$, $r_{size}$ and $r_{spacing}$ represent the Numerosity, Size and Spacing ratios between the two stimuli, and the corresponding $\beta$ coefficients represent the degree to which each orthogonal dimension affects discrimination performance. The $\beta_{Side}$ coefficient accounts for spatial response biases independent from stimuli properties, while the term $\gamma$ represents the guessing factor accounting for occasional random responses due to distraction. The individual guessing factor was estimated to minimize the deviance of the model and was set to 0.01.

***Projection Analysis.*** The direction of the discrimination vector defined by the coordinates $\beta_{Num}$, $\beta_{Size}$ and $\beta_{Spacing}$ represents what stimulus features are being mostly used to perform the discrimination, while the magnitude of the vector represents the participant's acuity in discriminating each feature. In the case of a strategy based exclusively on numerosity the discrimination vector will coincide with the *Numerosity* dimension, and the magnitude of the vector will be exactly $\beta_{Num}$. When choice is modulated by other dimensions, the contribution of each non-numerical feature is geometrically characterized by projecting the discrimination vector onto each dimension and statistically measure which one is the closest. Similarly, the most representative features can be quantified by measuring the angle between the discrimination vector and each candidate dimension (12). Multiple comparison tests were always corrected using the Bonferroni method. When the assumption of normality was violated, non-parametric tests (Wilcoxon signed rank) were performed.

***Deep Learning.*** Deep belief networks were first trained in a completely unsupervised way on the stimulus set containing all numbers from 5 to 32 (images were downscaled to 100x100 pixels for computational convenience). Deep networks were built as a stack of two Restricted Boltzmann Machines (RBMs) trained using contrastive divergence (46). Each RBM consisted of two layers of stochastic neurons, fully connected with symmetric weights and without self-connections (SI for details). Visual stimuli were provided as input by clamping the vectorized images on the visible neurons of the first RBM; the subsequent activation of hidden neurons constituted the model's internal representation of the stimulus. A decision layer was then trained by feeding the internal representations of two paired stimuli to a linear network with two output units implementing a binary classification task. We tested 12 different architectures (obtained by varying the number of hidden units in each layer) with 12 different random initializations of the connection weights, for a total of 144 networks. The source code of our simulations is available for download on the Open Science Framework, along with a copy of the trained networks that can be used to further test our model over different types of stimuli and experimental settings[4].

***Representational Similarity Analysis.*** Results reported in the main text refer to the architecture achieving the best numerosity discrimination performance, which had 1500 neurons in the first hidden layer and 1000 neurons in the second layer (control simulations showed that the overall findings hold also if we consider the complete set of architectures). The deep network was probed on a subset of test stimuli created by randomly selecting 10 instances of each combination of three Numerosity levels (7, 18, 28), three Size levels ($2.60 \times 10^5$, $6.55 \times 10^5$, $10.40 \times 10^5$) and 3 Spacing levels ($0.80 \times 10^7$, $2.02 \times 10^7$, $3.20 \times 10^7$), all equally spaced between the minimum and maximum values in the original stimulus set, for a total of 270 images. The activation patterns of the deepest hidden layer corresponding to instances with the same combination of features were averaged, resulting in 27 mean activation patterns. These mean activation patterns were then compared in order to build a Representational Dissimilarity Matrix (RDM), whose cells contained a number reflecting the dissimilarity (1 – Pearson correlation) between the internal representations associated to each combination of stimulus features. For comparison, categorical RDMs corresponding to all possible individual features were built by using as dissimilarity measure the difference between feature values on a log scale. The model RDMs were quantitatively compared to the categorical RDMs using Kendall's Tau alpha correlation, and their specific relatedness was statistically assessed using one-sided Wilcoxon signed rank tests, considering the simulated RDMs as reference and treating the categorical models as possible candidates. RSA was performed using a publicly available MATLAB toolbox (71).

---

[4] https://osf.io/j7dvc

## Acknowledgements

This work was supported by Cariparo Excellence Grant 2017 ("Numsense") to M.Z. and STARS Starting Grant 2018 ("Deepmath") to A.T. We also acknowledge the support of NVIDIA Corporation for the donation of a Titan Xp GPU used for this research.

# Supplementary Methods

## Stimulus space definition

The relationship between the three orthogonal dimensions *Numerosity*, *Size* and *Spacing* and the individual non-numerical features can be algebraically defined according to the following equations, where *n* is the number of items, *TSA* is the total surface area, *ISA* is the item surface area, *FA* is field area (approximation of convex hull) and *Spar* is sparsity (inverse of density). Note that, after log scaling the axes, the distance between stimulus points in the space is proportional to the ratios of their features, and the equations relating the orthogonal dimensions to the other non-numerical features are all linear equations (53).

$$log_2(n) = log_2\left(\frac{TSA}{ISA}\right) = log_2\left(\frac{FA}{Spar}\right)$$

$$log_2(Size) = log_2(TSA) + log_2(ISA)$$

$$log_2(Spacing) = log_2(FA) + log_2(Spar)$$

Individual features in terms of the three orthogonal dimensions:

$$TSA = \sqrt{Sz \cdot n}$$

$$ISA = \sqrt{\frac{Sz}{n}}$$

$$FA = \sqrt{Sp \cdot n}$$

$$Spar = \sqrt{\frac{Sp}{n}}$$

$$TP = 2\sqrt{\pi} \cdot Sz^{\frac{1}{4}} \cdot n^{\frac{3}{4}}$$

$$IP = 2\sqrt{\pi} \cdot Sz^{\frac{1}{4}} \cdot n^{-\frac{1}{4}}$$

$$Cov = \sqrt{\frac{Sz}{Sp}}$$

$$AC = \sqrt{Sz \cdot Sp}$$

Log of each feature in terms of log of the three orthogonal dimensions:

$$log_2(TSA) = \frac{1}{2}log_2(Sz) + \frac{1}{2}log_2(n)$$

$$log_2(ISA) = \frac{1}{2}log_2(Sz) - \frac{1}{2}log_2(n)$$

$$log_2(FA) = \frac{1}{2}log_2(Sp) + \frac{1}{2}log_2(n)$$

$$log_2(Spar) = \frac{1}{2}log_2(Sp) - \frac{1}{2}log_2(n)$$

$$log_2(TP) = log_2(2\sqrt{\pi}) + \frac{1}{4}log_2(Sz) + \frac{3}{4}log_2(n)$$

$$log_2(IP) = log_2(2\sqrt{\pi}) + \frac{1}{4}log_2(Sz) - \frac{1}{4}log_2(n)$$

$$log_2(Cov) = \frac{1}{2}log_2(Sz) - \frac{1}{2}log_2(Sp)$$

$$log_2(AC) = \frac{1}{2}log_2(Sz) + \frac{1}{2}log_2(Sp)$$

**Simulations details**

Deep belief networks were implemented as a stack of Restricted Boltzmann Machines (RBMs) (46, 72). The dynamics of each RBM was driven by an energy function $E$ that specifies which configurations of the neurons are more likely to occur by assigning them a probability value:

$$p(v,h) = \frac{e^{-E(v,h)}}{Z}$$

where $v$ and $h$ represent the visible and hidden neurons and $Z$ is the partition function. Since there are no connections within the same layer the energy function can be defined as:

$$E(v,h) = -b^T v - c^T h - h^T W v$$

where $W$ is the matrix of connections weights and $b$ and $c$ are the biases of visible and hidden neurons, respectively. RBMs were trained using 1-step contrastive divergence (73). For each training pattern, during the positive phase all visible neurons are clamped to the current pattern, and the activation of hidden neurons is computed by sampling from their conditional probability:

$$P(h|v) = \prod_{j=1}^{n} P(h_j|v)$$

where $n$ is the number of neurons in the hidden layer. Conditional activation probabilities for each single neuron are computed using the sigmoid logistic function:

$$P(h_j=1|v) = \frac{1}{1+e^{-b_j - \sum_{i=1}^{m} w_{ij} v_i}}$$

where $m$ is the number of neurons in the visible layer. During the negative phase, the activation of the hidden neurons corresponding to the clamped data pattern is similarly used to perform top-down reconstruction of the stimulus over the visible neurons. Connection weights were randomly initialized according to a Gaussian distribution with zero mean and standard deviation of 0.01. Learning hyperparameters were optimized by systematically varying the learning rate (best value 0.15), weight decay factor (best value 0.0001), momentum coefficient (best value 0.7) and mini-batch size (best value 100). Learning was performed using in-house source code optimized for graphic processing units (GPUs) (74).

The model was build using the two-layer architecture adopted in previous studies (51, 52) and explored possible variations by systematically changing the size of the first (H1) and second (H2) hidden layers, resulting in 12 combinations:

|    | N.1 | N.2 | N.3 | N.4 | N.5 | N.6 | N.7 | N.8 | N.9 | N.10 | N.11 | N.12 |
|----|-----|-----|-----|-----|-----|-----|-----|-----|-----|------|------|------|
| H1 | 500 | 500 | 500 | 500 | 1000| 1000| 1000| 1500| 1500| 1500 | 1500 | 1500 |
| H2 | 500 | 1000| 1500| 2000| 500 | 1000| 1500| 2000| 500 | 1000 | 1500 | 2000 |

The supervised read-out layer was trained using an efficient implementation of linear associative learning formalized according to the pseudoinverse method (45, 75). The read-out layer simultaneously received the pattern of activity elicited by two different input images and was trained to assess which of the two contained the larger numerosity, as in (32, 52). 15200 training pairs were used to train the read-out, while a different combination of 15200 image pairs was used to test it. It should be noted that the model's test set thus more systematically covered the stimulus space, compared to the human's test set that only included 300 image pairs. It should also be noted that the classifier weights were not further tuned during testing, thus our model does not take into account practice effects that might have occurred during human testing.

**Representational Similarity Analysis (RSA)**

RSA was carried out using the methods and toolbox described in (54, 71). The activation patterns of the deepest hidden layer in response to the selected stimuli were extracted and averaged across instances of the same combination of features, resulting in 27 mean activation patterns. These mean activation patterns were then compared, computing a Representational Dissimilarity Matrix (RDM). The latter is a symmetric matrix containing pairwise dissimilarity measures between the internal representations associated to each combination of stimulus features. Given the number of stimuli, our resulting RDMs were 27x27 matrices encoding dissimilarity as 1 – Pearson correlation between each pair of representations. For comparison, categorical models for every individual stimulus feature were built by computing artificial RDMs encoding dissimilarity between the stimuli as their pairwise difference in that feature on a logarithmic scale. Categorical models were created for numerosity, field area, total surface area, item surface area, sparsity, convex hull, total perimeter and item perimeter (Fig. 4A).

The categorical RDMs were quantitatively compared to the model RDMs using Kendall's rank correlation, and their specific relatedness was statistically assessed, separately for Young and Mature models, by computing the correlation between each categorical RDM and the separate instances of the model RDMs and performing a one-sided Wilcoxon signed rank test against the hypothesis of no correlation (Fig. 4C). We also performed the same analysis using alternative tests available in the RSA toolbox (i.e., randomization and bootstrapping), obtaining similar results. Multiple comparisons were corrected using False Discovery Rate, setting to 0.01 the expected proportion of categorical RDMs falsely declared significant among all candidate RDMs declared significant. Pairwise comparisons between categorical RDMs were also conducted to assess possible differences in their relatedness to the model RDMs, separately for Young and Mature networks: for each pair of categorical RDMs, two-sided Wilcoxon signed rank tests were performed comparing their correlation with the model RDMs, against the null hypothesis of an equal correlation with the model (Fig. S2). Correction for multiple comparisons was carried out using FDR with threshold 0.01. RDMs shown in Fig. 4A are colored according to the standard toolbox colormap "blue-cyan-gray-red-yellow".

**t-Distributed Stochastic Neighbor Embedding (t-SNE)**

t-SNE projects high-dimensional objects in a two-dimensional space, with the goal of mapping with high probability similar objects into nearby points and dissimilar objects into distant points. The algorithm comprises two main stages. First, it constructs a probability distribution over pairs of high-dimensional objects in such a way that similar objects have a high probability of being picked, whilst dissimilar points have a very low probability of being picked. Second, it defines a similar probability distribution over the points in the low-dimensional map, and it minimizes the Kullback–Leibler divergence between the two distributions with respect to the locations of the points in the map (56).

In our analysis, we considered the best performing deep learning architecture and we used as high-dimensional objects the activation of the neurons in the second hidden layer corresponding to each stimulus, divided according to four different congruency conditions: 1) Numerosity, Size and Spacing all congruent; 2) Numerosity and Size congruent, Spacing incongruent; 3) Numerosity and Spacing congruent, Size incongruent; 4) Size and Spacing congruent, Numerosity incongruent. As in common practice, the t-SNE cluster analysis was performed on the first 20 components extracted from a principal component analysis over the activation's matrix. Distances between data points were measured using Euclidian distance and the perplexity parameter was set to 100. In the resulting two-dimensional plot (see Fig. S3) the display numerosity is represented using different colors, according to a standard JET colormap.

**Simultaneous *vs.* sequential presentation of the stimuli**

All subjects involved in the study also carried out a sequential version of the numerosity discrimination task. A fixation cross was presented for 500 ms; then a first cloud of dots appeared on the left or on the right of fixation for 250 ms, followed by a mask lasting for 150 ms and then by a second cloud of dots on the opposite side of the fixation, again for 250 ms. As for the simultaneous task, a final mask of 150 ms was then presented. The side of appearance of the first array was randomized. A black response screen was then presented until subject's response. The inter-trial interval ranged between 1250 and 1750 ms.

Discrimination accuracy was well above chance (86.13%, range: 65.85-93%). Comparing the performance in the two versions of the numerosity comparison task, a two-by-two repeated measures ANOVA revealed a significant difference in accuracy between the simultaneous and sequential tasks ($F(1,36) = 9.23$, $p < .01$, $\eta^2 = 0.2$) but no significant effect of task order and no interaction between order presentation and task variant. The GLM fit for the sequential task was significant at individual subject level (mean adjusted $R^2 = .60$, mean chi-square value = 211.17, all $p < .001$). Coefficient fits for each orthogonal feature were significantly different from zero for *βNum* ($t(37) = 21.45$, $p < .001$), *βSize* ($t(37) = -2.71$, $p < .05$) and *βSpacing* ($t(37) = 3.67$, $p < .001$). The projection analysis comparing *βNum* to coefficient weights of all the other features revealed that *βNum* was significantly larger than all other projections (all ts > 9.53, all ps < .001). Coherently with the projection results, in the angle analysis Numerosity resulted the closest dimension to the discrimination vector (8.71 deg), followed by Total Perimeter (22.21 deg), with a significant angle between these two closest features ($z = 5.23$, $p <$

.001). Comparing the two task variants, the angle between the discrimination vector and the Numerosity dimension resulted statistically larger in the simultaneous task than in the sequential task ($t = 2.14$, $p < .05$, $d = 0.35$), whose discrimination vector resulted instead significantly farther from the closest non-numerical feature (Total perimeter) than in the simultaneous task ($t = 4.09$, $p < .001$, $d = 0.66$).

Note that our simulations focused on the data collected with the simultaneous task, because the latter is more consistent with the way the comparison task is implemented in the model. Indeed, to carry out the numerosity comparison the decision layer receives as input the internal representation of two stimuli at the same time.

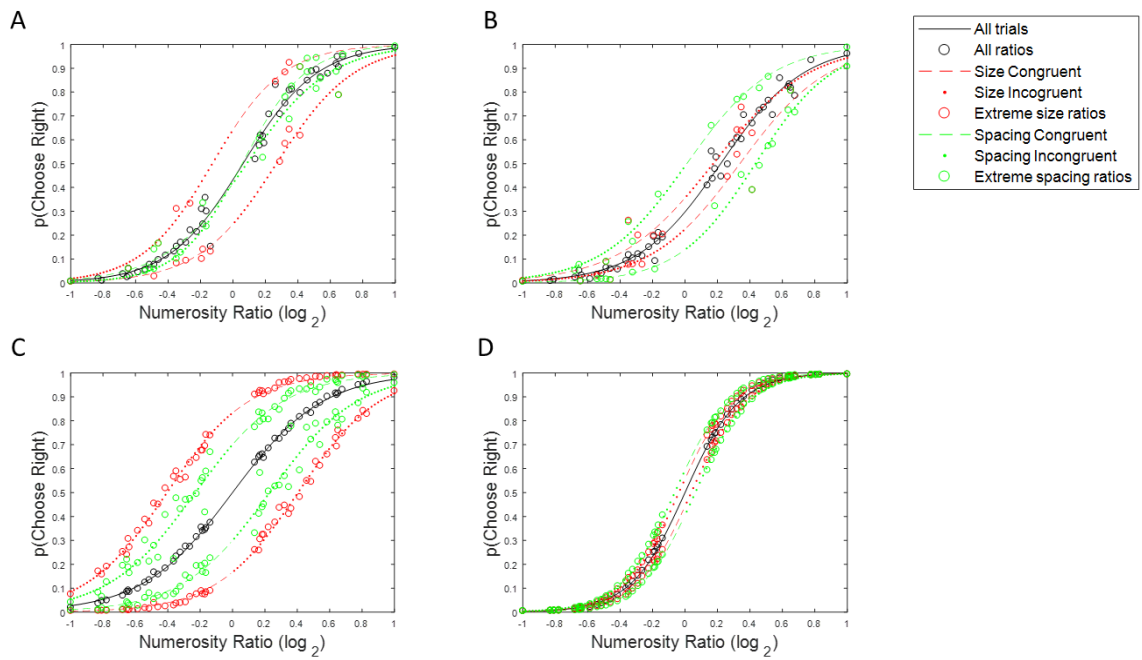

**Figure S1.** Model fit for two participants (A-B), one Young deep network (C) and one Mature deep network (D). The probability of choosing the right stimulus array is modeled as a function of the log of Numerosity, Size and Spacing ratios. Black lines indicate the model fit for all data (black circles). Red color shows data and model fits for the trials with extreme Size ratio, while green color shows data and model fit for trials with great Spacing ratio. The influence of Size and Spacing are determined by the offset between the colored lines and the original model fit, with a positive or negative effect on performance depending on the congruency between non-numerical features and number: dashed lines indicate that Size or Spacing were congruent with numerosity, while dotted lines indicate incongruent trials.

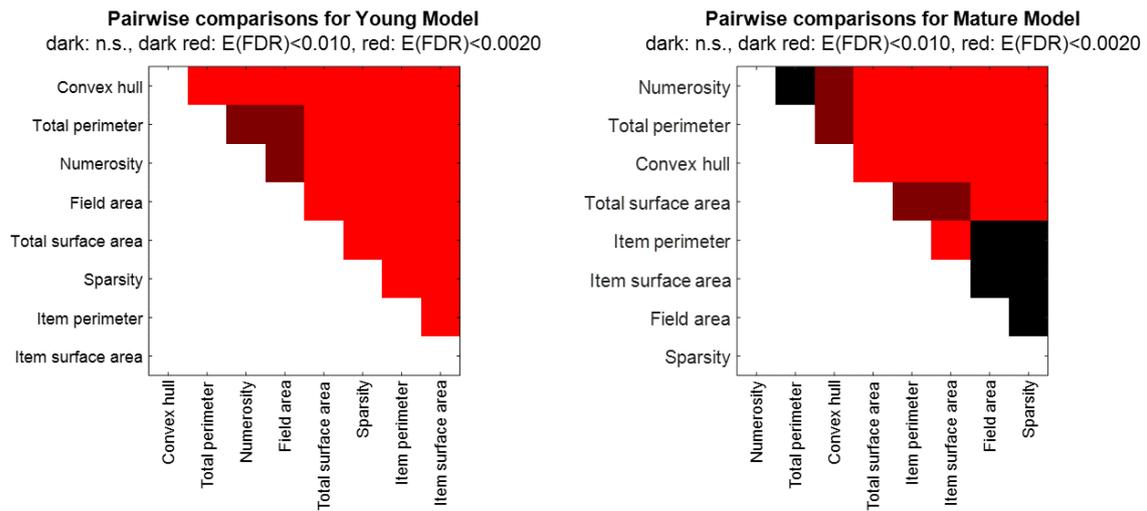

**Figure S2.** Pairwise comparisons between all candidate models in their relatedness to the reference RDM. The matrix shows the results of two-sided Wilcoxon signed rank tests for each pair of categorical RDM. Significance is based on different FDR thresholds, encoded using different levels of red: black indicates nonsignificant comparisons. CH: Convex Hull; TP: Total Perimeter; FA: Field Area; TSA: Total Surface Area; Spar: Sparsity; IP: Item Perimeter; ISA: Item Surface Area.

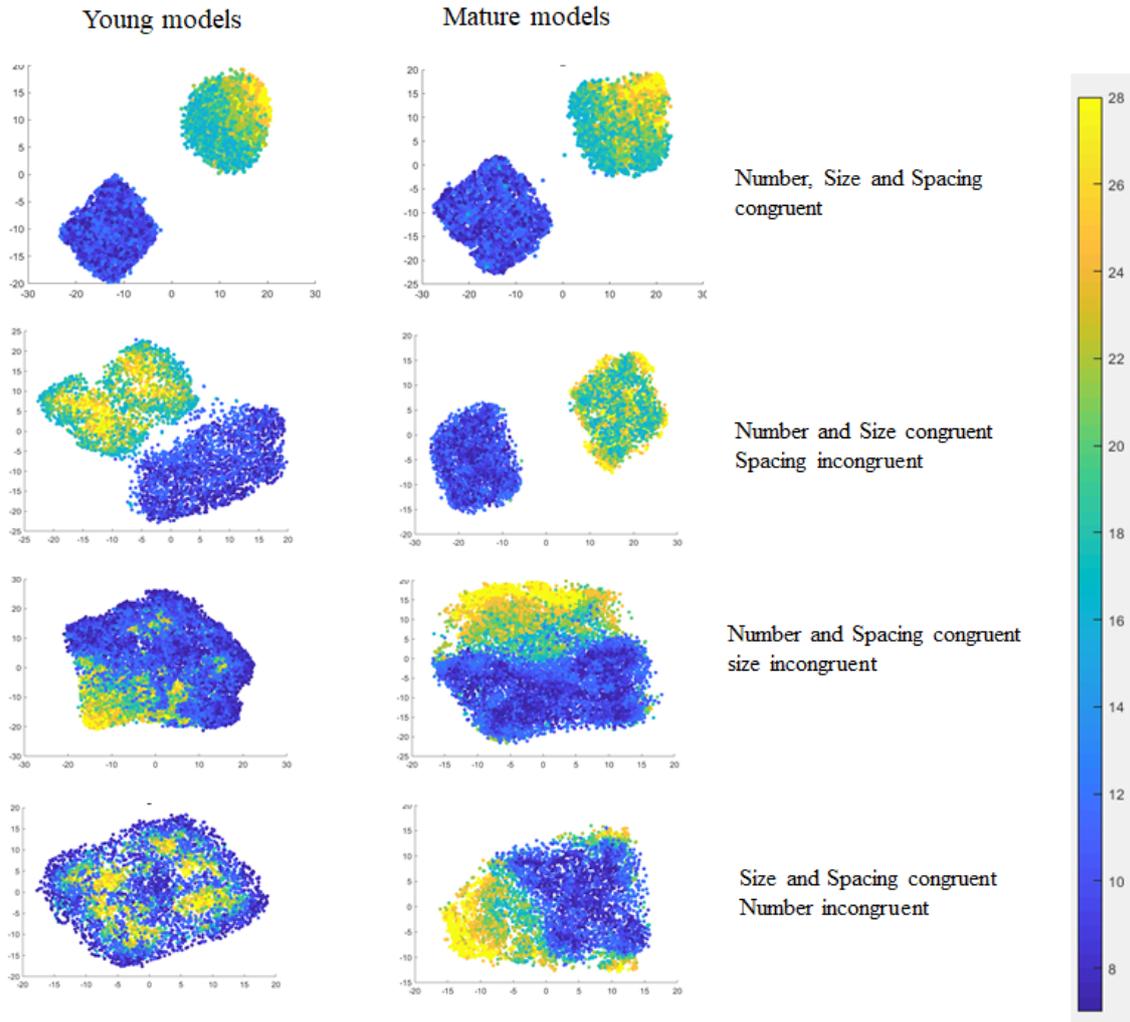

**Figure S3.** Manifold projection using t-SNE. Stimuli with a small or large numerosity (respectively in the ranges 7:12 and 16:28) were first selected from the complete image data set. In the top panels, Numerosity, Size and Spacing are all congruent, which means that images with a small number of dots also have low Spacing and Size values. In the second-row panels, Numerosity and Size are congruent, but Spacing is not. In the third-row panels, Numerosity and Spacing are congruent, but Size is not. In the bottom panels, both Spacing and Size are incongruent with Numerosity. Results show that in the Mature model the representations are mostly clustered, with a distance gradient often proportional to number. In the Young model, when number is incongruent with Size the clustering almost disappears, especially when also Spacing is incongruent.